# A Robust Approach for the Decomposition of High-Energy-Consuming Industrial Loads with Deep Learning

Jia Cui [a,*], Yonghui Jin [a], Renzhe Yu [b], Martin Onyeka Okoye [a], Yang Li [d*], Junyou Yang [a], Shunjiang Wang [c*]

a The School of Electrical Engineering, Shenyang University of Technology, Shenyang 110870, Liaoning Province, China.

b Renzhe Yu is with Ulanqab Electric Power Bureau, Ulanqab 012000, Ulanqab, China.

c Shunjiang Wang is with State Grid Liaoning Electric Power Supply Co, Ltd., Shenyang 110006, Liaoning Province, China.

d Yang. Li is with the School of Electrical Engineering, Northeast Electric Power University, Jilin 132012, China.

**Abstract**：The knowledge of the users' electricity consumption pattern is an important coordinating mechanism between the utility company and the electricity consumers in terms of key decision makings. The load decomposition is therefore crucial to reveal the underlying relationship between the load consumption and its characteristics. However, load decomposition is conventionally performed on the residential and commercial loads, and adequate consideration has not been given to the high-energy-consuming industrial loads leading to inefficient results. This paper thus focuses on the load decomposition of the industrial park loads (IPL). The commonly used parameters in a conventional method are however inapplicable in high-energy-consuming industrial loads. Therefore, a more robust approach is developed comprising a three-algorithm model to achieve this goal on the IPL. First, the improved variational mode decomposition (IVMD) algorithm is introduced to denoise the training data of the IPL and improve its stability. Secondly, the convolutional neural network (CNN) and simple recurrent units (SRU) joint algorithms are used to achieve a non-intrusive and non-invasive decomposition process of the IPL using a double-layer deep learning network based on the IPL characteristics. Specifically, CNN is used to extract the IPL data characteristics while the improved long and short-term memory (LSTM) network, SRU, is adopted to develop the decomposition model and further train the load data. Through the robust decomposition process, the underlying relationship in the load consumption is extracted. The results obtained from the numerical examples show that this approach outperforms the state-of-the-art in the conventional decomposition process.

*Corresponding author
E-mail address: cuijiayouxiang@sut.edu.cn (J. Cui)

# 1. Introduction

With the development of power and Internet of things, edge cloud (Golpîra and Bahramara, 2020) plays a more and more important role in the smart grid. As a specific manifestation of edge cloud, non-intrusive load monitoring (NILM) has attracted the attention of many scholars. Compared with intrusive load monitoring (ILM) (Rana et al, 2020), non-intrusive load monitoring only needs to use the load port monitoring information for "soft computing" instead of "hard measurement" at the equipment, which is economical and convenient. Most of the existing researches are mostly the non-intrusive load monitoring based on mathematical optimization and pattern recognition, which have a good effect on obtaining internal data of users. In the smart grid environment, demand response (DR) program has attracted considerable attention for smart buildings (Cui et al., 2022) and industrial electricity consumers (Cui and Zhou, 2018). Recognition of flexible loads is essential for managing energy systems of factories and industrial parks. However, the researchers are mostly focused on residential and commercial loads (Verma et al., 2021; Azizi et al., 2021; Meier and Cautley, 2021; Brucke et al., 2021). The non-intrusive decomposition and identification for high energy-consuming industrial loads lack in-depth researches, and the researches on industrial park loads (IPL) are even less. Many challenges in developing a load aggregator model decomposition system for industrial parks have rarely been studied.

Industrial loads may be influenced by many factors, such as scheduled processes and work shifts, which are uncommon or unnecessary in the perception of classical load models (Wang et al., 2018). The industrial load operating all day would most likely have a different load pattern profile from those only operating during the weekdays. A manually controlled machine may have a different power profile from a furnace or an automated machine. Additionally, the power may or may not be responsive to the weather. Even within the same factory, the equipment load levels may be different from each other. Thus, the specific calendar, weather, and electrical variables must be identified and investigated before building accurate load decomposition models. Moreover, the features of load patterns are significantly different from residential and commercial loads, so some commonly used variables in classical load decomposition models may become meaningless for decomposing industrial park loads aggregator data. At present, there are few studies on non-intrusive load monitoring research around the world, because some problems lead to unsatisfactory results when decomposing the total load curve of industrial parks. For example, the difficulty in collecting operation information of high energy-consuming industrial load equipment leads to the lack of relevant data sets, larger equipment power consumption, less internal power equipment interruption model effect, and the diversification of industrial park load.

Data preprocessing is the first step of non-invasive load monitoring. For the preprocessing of industrial park loads data, where the data set of regional industrial load measurement is different from the REDD used in the existing research results, it should be processed before decomposing the load data to remove the internal factors that affect the accuracy of non-invasive load decomposition (NILD), such as noise and error data. Ch'ien et al propose that the empirical modal decomposition (EMD) algorithm (Ch'ien et al., 2020) is used for signal processing. The interference signal is identified and suppressed through joint analysis of the intrinsic mode function (IMF) of the double-channel signals. However, the mode aliasing and edge effect of EMD seriously affect the accuracy of noise reduction. Xie et al propose a white noise suppression method based on short-time singular value decomposition (STSVD) (Xie et al., 2019). The short-time data window is used to intercept the noisy partial discharge signal segments for singular value decomposition, which can improve the execution efficiency of the algorithm. However, it is necessary to perform singular value decomposition and reconstruction for each signal intercepted segment. When data is large, the execution efficiency of this method would be reduced. The empirical wavelet decomposition method (Arnold et al., 2018) is used to achieve data denoising, combined with empirical modal decomposition and traditional wavelet denoising method. The time-frequency hybrid noise reduction method (Takada and Satoh, 2018) divides the problem of signal noise reduction into two steps that are signal frequency-domain noise reduction and carrier time-domain noise reduction. The effect of noise reduction is always limited at the point of signal discontinuity or in the case of the useful signal overlaps noise frequency band in these studies, which would result in slowing down the speed of denoising as the signal strength increases. Therefore, it is necessary to remove the noise by decomposing the signal before being reconstructed. Else, the distortion would appear at the peak of reconstructed data after partial noise reduction and lead to a reduction in the accuracy of data decomposition.

The existing non-intrusive load monitoring methods mainly include traditional methods, statistical learning methods and deep learning methods. The traditional methods realize load decomposition by marking some load marks, and use cluster analysis algorithm to identify various types of electrical equipment. Qureshi et al propose an event-based blind disaggregation algorithm (Qureshi et al., 2021) that uses Gaussian mixture models (GMM) for clustering to automatically detect two-state appliances from the aggregate data. Bayesian information criterion (BIC) is used to determine the number of clusters. Unsupervised algorithms are suitable for scenarios where label data are not easy to obtain, but their accuracy is usually lower than supervised algorithms. Statistical learning methods decompose load by establishing a probability model of load behavior. Wu et al combine an adaptive density peak clustering (ADPC) and factorial hidden Markov model (FHMM) to create an adaptive density peak clustering-factorial hidden Markov model (ADPC-FHMM) (Wu et al.,2021), that automatically determine the working states of appliance based on its power consumption and reduce the dependence of prior information. The disadvantage of this method is that it is difficult to solve the decomposition of devices with multiple state changes. Moreover, the model is complex and the convergence speed of iteration is slow. Recently, deep learning methods have been applied to non-intrusive load monitoring due to their powerful feature representation learning capabilities. Deep learning methods have shown great promise for non-intrusive load monitoring. Convolutional neural network (CNN) and long and short-term memory (LSTM) network (Zhou et al., 2021) are used to extract load labels of a single device and decompose the aggregated loads power. Subash et al use deep neural network (Subash et al.,2021) to decompose the energy use of household equipment. In the presence of vapor loads of households，the gated recurrent unit (GRU) has achieved good decomposition effect. The above methods only consider the internal characteristics of the load, and ignore the significant impact of non- electrical factors on the load behavior, such as weather, the specific calendar. At the same time, the existing non-intrusive load monitoring method relies on the state of the pre-order equipment to a great extent for the decomposition process, so it usually appears that the data precision of the follow-up training is affected by the lack of training precision at a certain point in the training set.

Given the above research shortcomings, this paper proposes an aggregation model recognition method for industrial parks that integrates data processing and non-invasive decomposition. Firstly, it is necessary to retain the data containing noise due to the limited amount of data and then preprocess them. Therefore, the improved variational mode decomposition (IVMD) algorithm is introduced in this paper to decompose data, using the time-frequency conversion function of variational mode decomposition (VMD) to convert it into the frequency-domain signal, which can distinguish the noises according to the signal frequency and eliminate them. Secondly, CNN is adopted to extract characteristics from preprocessed data for high energy-consuming industrial loads. Furthermore, a two-tier network training model is formed by using SRU to decompose the high energy-consuming industrial park loads with high precision. Finally, the dataset of the measured load of an industrial park in Liaoning Province is analyzed. Compared with the existing non-invasive load decomposition algorithm, the results prove that the proposed algorithm is effective and adaptive, and improves the identification accuracy of load aggregator for industrial park aggregation model.

This paper brings a new overlapping problem to the industrial load decomposition literature. The problem is significant because of the inhomogeneous inputs that must be considered in the model building. The contributions of this paper are summarized as follows:

(1) Firstly, the load curve curvature index (LCC) is proposed to improve the VMD algorithm. By quantifying the mean value of the instantaneous frequency of the eigenmode after improved variational mode decomposition (IVMD), the optimal value of decomposition mode number K is obtained. The optimization of the K-value would help VMD reflect the variation characteristics of the load curve better and more accurately remove noise from IPL data.

(2) Because of the characteristics of fewer interrupts and high power of IPL data, the timestamp and coding methods used in the traditional methods are incapable due to its high dependence on high-quality history data. CNN is used to extract the characteristics of preprocessed IPL data and other real-time relative data, and SRU is used to train them to form an enhanced two-layer network training model. Through CNN feature extraction, the internal feature extraction ability of SRU is improved. Moreover, the problem of over-dependence on the previous state in the training process of the conventional decomposition algorithm is solved resulting in a higher decomposition accuracy of IPL data.

(3) This paper fills a gap in the literature by studying an elaborate decomposition framework for load data in industrial parks which integrates IPL data preprocessing and decomposition. Compared with residential and commercial load, industrial park load is characterized by more influenced factors, lower start-stop frequency, greater flexibility potential, the difficulty of data acquisition, and poor raw data quality. Conventional methods rarely combine data preprocessing and decomposition, and the perception is not transparent enough. In this paper, the denoising ability of the proposed algorithm based on IVMD combined with CNN and SRU (CNN-SRU) double network is improved significantly. The denoising effect of IPL data is considered to improve the decomposition accuracy and generalization ability. To make the comparisons, two common algorithms of deep learning and mathematical optimization are selected. The attention-seq2seq represents the deep learning algorithm, and the aided linear integer programming (ALIP) represents the mathematical optimization algorithm. The effectiveness of the algorithm is verified by the actual load data of an industrial park in Liaoning province, China.

The rest of the paper is organized as follows: Section 2 presents a method of determining an optimal K value using eigenmode instantaneous frequency for VMD. Section 3 describes the load data decomposition method of the industrial park based on CNN-SRU. Section 4 builds a load data decomposition framework based on the IVMD denoising algorithm and CNN-SRU to provide theoretical support for the case analysis of Section 5. Finally, a conclusion is provided in Section 6.

## 2. Noise Reduction of Industrial Load Data Based on Improved VMD

*2.1. Improved VMD algorithms*

As a novel complete non-recursive modal variational method, the VMD determines the bandwidth and center frequency by finding the optimal solution of the variational model within the variational framework iteratively. The initial signal is decomposed into K eigenmodes with different frequencies, that is, each mode has a different frequency when decomposed fully. However, the frequencies may be confused and overlapped when over decomposed. Therefore, in this paper, K value is optimized by combining instantaneous frequency and curvature.

The IVMD algorithm flow is as follows:

2.1.1. Initialization of K value

The VMD variational problem be described as seeking K modal functions $u_k(t)$ to minimize the sum of the estimated bandwidth of each mode. The constraint is that the sum of the modals is equal to the input signal $f_{IPL}$, that is, high energy-consuming industrial load data. The K value needs to be initialized by the first decomposition.

(a) The constrained variational problem is constructed by Hilbert transform as follows:

$$\begin{cases} \min_{\{u_k\},\{w_k\}} \left\{ \sum_k \left\| \partial_t \left[ (\delta(t) + \frac{j}{\pi t}) * u_k(t) \right] e^{-j\omega t} \right\|_2^2 \right\} \\ \text{s.t.} \sum_k u_k(t) = f_{IPL} \end{cases} \quad (1)$$

where $e^{(-j\omega t)}$ is the phase description of the center frequency in the complex plane, $\{u_k\} = \{u_1, \cdots, u_K\}$, $\{\omega_k\} = \{\omega_1, \cdots, \omega_K\}$, $\sum_k := \sum_{k=1}^{K}$.

(b) By introducing the quadratic penalty factor $\alpha$ and the Lagrangian multiplication operator $\lambda(t)$, the constrained variational problem is transformed into the unconstrained variational problem. The extended Lagrange expression is as follows:

$$L(\{u_k\},\{\omega_k\},\lambda) := \alpha \sum_k \left\| \partial_t \left[ (\delta(t) + \frac{j}{\pi t}) * u_k(t) \right] e^{-j\omega_k t} \right\|_2^2 + \left\| f_{IPL}(t) - \sum_k u_k(t) \right\|_2^2 + \langle \lambda(t), f_{IPL}(t) - \sum_k u_k(t) \rangle \quad (2)$$

(c) Alternative direction method of multipliers (ADMM) is adopted in VMD to solve the above variational problem, in which the value of $u_k^{n+1}$ can be expressed as:

$$\hat{u}_k^{n+1} = \arg\min_{\hat{u}_k, u_k \in X} \left\{ \alpha \left\| \partial_t \left[ (\delta(t) + \frac{j}{\pi t}) * u_k(t) \right] e^{-j\omega_k t} \right\|_2^2 + \left\| f_{IPL}(t) - \sum_i u_i(t) + \frac{\lambda(t)}{2} \right\|_2^2 \right\} \quad (3)$$

where $\omega_k$ is equal to $\omega_k^{n+1}$; $\sum_i u_i(t)$ is equal to $\sum_{i \neq k} u_i(t)^{n+1}$.

By using the Parseval/Plancherel Fourier equidistant transform, formula (3) is transformed into frequency domain:

$$\hat{u}_k^{n+1} = \arg\min_{\hat{u}_k, u_k \in X} \left\{ \alpha \left\| j\omega[1 + sgn(\omega + \omega_k)\hat{u}(\omega + \omega_k)] \right\|_2^2 + \left\| \hat{f}_{IPL}(\omega) - \sum_i \hat{u}_i(\omega) + \frac{\hat{\lambda}(\omega)}{2} \right\|_2^2 \right\} \quad (4)$$

The $\omega$ of the first term is replaced with $\omega - \omega_k$ (Dragomiretskiy and Zosso, 2014):

$$\hat{u}_k^{n+1} = \arg\min_{\hat{u}_k, u_k \in X} \left\{ \alpha \left\| j(\omega - \omega_k)[1 + sgn(\omega)\hat{u}(\omega)] \right\|_2^2 + \left\| \hat{f}_{IPL}(\omega) - \sum_i \hat{u}_i(\omega) + \frac{\hat{\lambda}(\omega)}{2} \right\|_2^2 \right\} \quad (5)$$

The formula (5) is converted into the form of non-negative frequency interval integrals:

$$\hat{u}_k^{n+1} = \arg\min_{u_k, u_k \in X} \left\{ \int_0^\infty 4\alpha(\omega - \omega_k)^2 |\hat{u}_k(\omega)|^2 + 2 \left| \hat{f}_{IPL}(\omega) - \sum_i \hat{u}_i(\omega) + \frac{\hat{\lambda}(\omega)}{2} \right|^2 d\omega \right\} \quad (6)$$

At the same time, the solution of the quadratic optimization problem is as follows:

$$\hat{u}_k^{n+1}(\omega) = \frac{\hat{f}_{IPL}(\omega) - \sum_{i \neq k} \hat{u}_i(\omega) + \frac{\hat{\lambda}(\omega)}{2}}{1 + 2\alpha(\omega - \omega_k)^2} \quad (7)$$

$$\omega_k^{n+1} = \frac{\int_0^\infty \omega |\hat{u}_k(\omega)|^2 d\omega}{\int_0^\infty |\hat{u}_k(\omega)|^2 d\omega} \tag{8}$$

where $\hat{u}_k^{n+1}(\omega)$ is equivalent to the Wiener filter of the current residual $\hat{f}_{IPL}(\omega) - \sum_{i \neq k} \hat{u}_i(\omega)$; $\omega_k^{n+1}$ is the gravity center of the modal function power spectrum.

2.1.2. Calculation of the optimal K value

The frequency $\omega_k$ obtained from formula (8) is taken as the input data. And the instantaneous frequency mean value (Cui et al.,2019) of each component is calculated. The curvature is used to analyze the mean value of instantaneous frequency and judge the change in the mean value of instantaneous frequency.

The curvature of the instantaneous frequency mean curve $K_{IPL}$ is:

$$K_{IPL} = \frac{1}{\rho} = \frac{|\bar{\omega}_k''|}{(1 + \bar{\omega}_k'^2)^{3/2}} \tag{9}$$

where $\rho$ is the radius of curvature; $\bar{\omega}_k$ is the mean value of instantaneous frequency. The optimal K value is determined by the change in $K_{IPL}$, the second decomposition is realized by inputting formula (9), and then the optimal decomposition result is obtained.

2.1.3. The calculation process of the IVMD algorithm

Step1. Initializing $\{\hat{u}_k^1\}$, $\{\omega_k^1\}$, $\{\hat{\lambda}_k^1\}$, and $n$;
Step2. If the discrimination accuracy $e > 0$, the iteration would be stopped when $\left[\sum_k \|\hat{u}_k^{n+1} - \hat{u}_k^n\|_2^2 / \|\hat{u}_k^n\|_2^2\right] < e$;
Step3. Taking $\omega_k$ as input data and calculating the modal instantaneous frequency mean curvature;
Step4. Updating the value of K and repeat the step1- step3.

*2.2. Algorithm validity verification*

To verify the effectiveness of IVMD algorithm for IPL data noise reduction, textile enterprise load data is selected as the input data, and then 96 points of load data are collected in an interval of 15 minutes to fully retain the load characteristics. The operation results of IVMD will be compared with the wavelet packet (WP) and wavelet packet-singular value decomposition (WP-SVD).

Firstly, the IPL data with noise is used as input data, and the K value is determined by calculating the mean curvature of instantaneous frequency. The mean value of the instantaneous frequency of the decomposition mode in different conditions is shown in Fig. 1(a). The curvature is used for quantitative analysis to eliminate the influence of subjective factors, and the curvature of the instantaneous frequency mean curve corresponding to different K values is shown in Fig. 1(b).

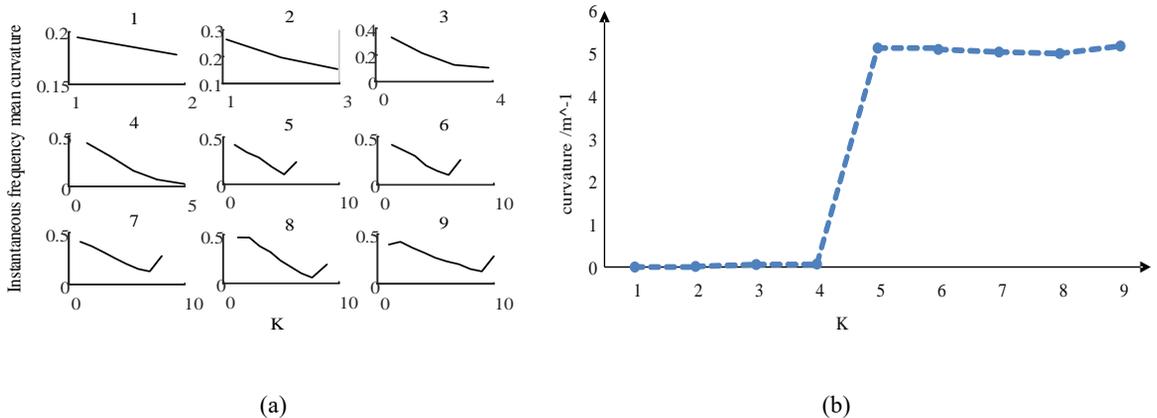

(a)  (b)

**Fig. 1.** Instantaneous frequency mean (a) Mean value (b) Curvature quantitative analysis

It can be seen from Fig. 1(a) that the trend of the instantaneous frequency curve changes significantly when the K value is greater than 4. It can be seen from Fig. 1(b) that the curvature of the frequency of each eigenmode changes significantly when the value of K is greater than 4, thus, the optimal K value is 4. The input parameters of the IVMD algorithm are determined as shown in Table 1.

**Table 1.** IVMD algorithm parameters

| Parameter | Bandwidth limit | Noise tolerance | Modal number | First center frequency update times | Initial center frequency | Termination condition |
|---|---|---|---|---|---|---|
| Value | 2000 | 0.001 | 4 | 0 | 1 | 1e-7 |

The part operation results of the IVMD and three noise reduction methods are shown in Fig. 2 and Fig. 3.

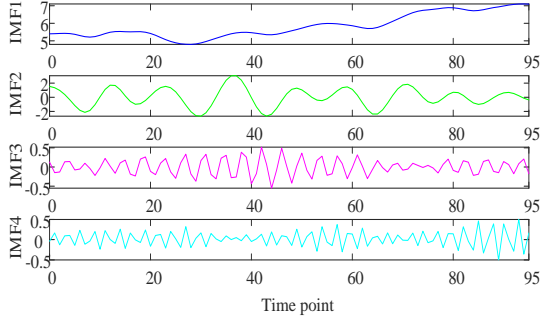 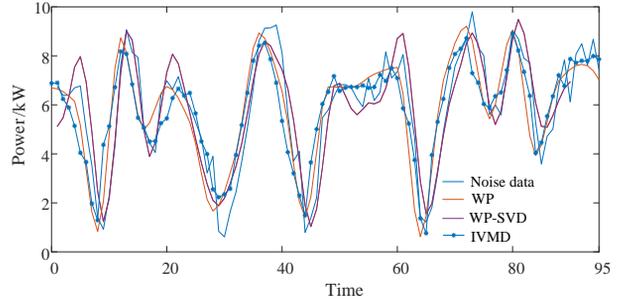

**Fig. 2.** IVMD operation results　　　　**Fig. 3.** Comparison of noise reduction effect of the algorithm

It can be seen from Fig. 3 that IVMD reduces the noise best, and retains the characteristics of initial load data effectively. While, the noise reduction effect of WP and WP-SVD is relatively poor, which deviates from the basic trend of the initial load data.

To test the advantages of the IVMD algorithm in data noise reduction more intuitively, the signal-to-noise ratios (SNR) is set to 20, 30 and 40 respectively by changing the noise intensity. The correlation coefficient (CC) (Cui et al.,2019) is used as the evaluation index, and the results are shown in Fig. 4.

$$CC = \frac{\sum_{i=1}^{n}(x_i - \bar{x})(y_i - \bar{y})}{\sqrt{\sum_{i=1}^{n}(x_i - \bar{x})^2}\sqrt{\sum_{i=1}^{n}(y_i - \bar{y})^2}} \tag{10}$$

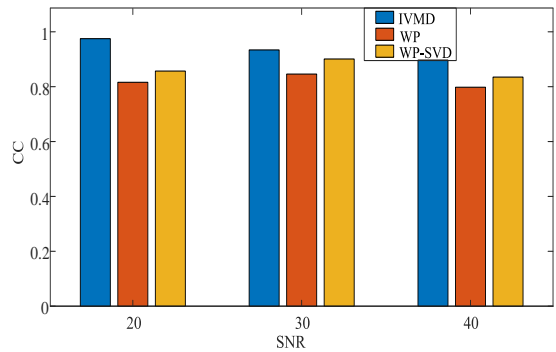

**Fig. 4.** Comparison of noise reduction results with different signal-to-noise ratios

It can be seen from Fig.4 that the effect of noise reduction of the IVMD algorithm is much better than the other two algorithms for IPL data with different signal-to-noise ratios. The correlation coefficient remains above 0.94 in textile enterprise results, which fully maintains the characteristics of load data. Compared with WP and WP-SVD, noise reduction accuracy is improved by 19.49% and 13.77%, respectively. Also, the denoising performance is not stable when the signal-to-noise ratio changes for the other two denoising algorithms.

## 3. Two-Tier Network Building Based on Improved Deep Learning

The load data quality for network training plays an important role in the application of deep learning algorithms to improve the NILM. For example, the widely used REDD dataset is obtained through long-term monitoring and modification. Therefore, it is an ideal and feasible method to generate a large amount of training data after extracting typical IPL load characteristics by obtaining typical IPL load short-term operation data. In this paper, CNN is utilized to process the collected high energy-consuming industrial load data. Then, the data is trained based on the memory characteristics of SRU and the parallel processing advantage. Then, the training model is built according to the IPL characteristics in the time domain to improve the training speed and accuracy.

*3.1. The multi-port industrial park data characteristic extraction by CNN network*

CNN is an efficient recognition method that has been developed in recent years and has attracted wide attention. Hubel and Wiesel found that the unique network structure can reduce the complexity of the feedback neural network effectively when they studied the neurons used for local sensitivity and direction selection in the cerebral cortex of cats, and hence, they proposed the CNN. However, CNN is widely used in the field of image processing, and so on, but few studies have been applied to the field of power systems. In this paper, CNN is used to process high energy-consuming industrial load data. Each pixel information of different types of high energy-consuming industrial load data is regarded as the input thereby extracting its features. The basic process of CNN is as follows:

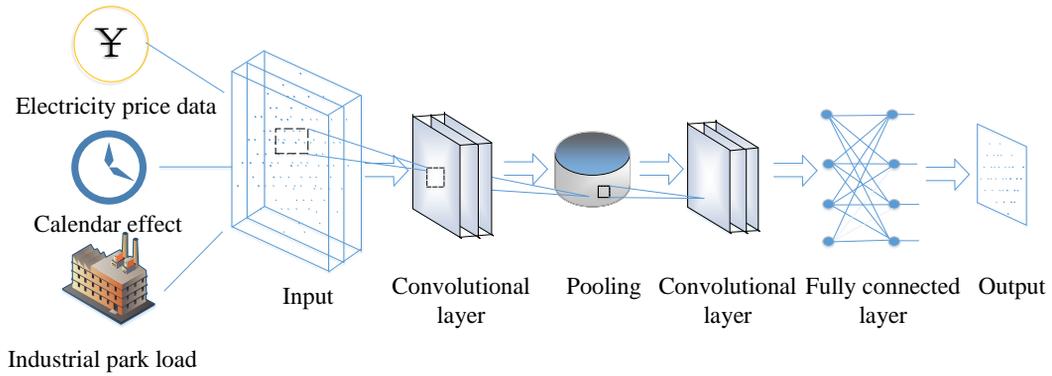

**Fig. 5.** CNN training process

3.1.1. Convolution layer

The main function of the convolution layer is to convolute the convolution kernel with the input data of the upper layer, and the output eigenvalue is used as the input of the lower layer. The input data used contains two main influence elements, which include the electricity price and calendar effect. The production of the enterprise is greatly affected by the electricity price in the same period (Sharma et al.,2018). Also, the electricity price directly affects the product cost of the enterprise. Furthermore, the calendar effect is selected as the input data because there is a big difference in electricity consumption between the usual time and the typical time. For example, the factory stops production on typical holidays in China. Thus, the input matrix is:

$$x = \begin{bmatrix} x_{e1} & \cdots & x_j \\ x_{c1} & \cdots & x_j \\ x_{IPL\ 1} & \cdots & x_{IPL\ j} \end{bmatrix} \quad (11)$$

where $x$ is the input matrix, $x_e$ is the electricity price data sequence, $x_c$ is the time sequence data, and $x_{IPL}$ is the industrial load sequence data.

The output eigenvalue y can be expressed as:

$$y_{i,j} = \sum_{p=1}^{P}\sum_{q=1}^{Q} x_{i+p-1, j+q-1} k_{p,q} \quad \begin{matrix} i=1,2,\cdots I \\ j=1,2,\cdots J \end{matrix} \quad (12)$$

where $y_{i,j}$ is the output matrix after convolution of the input data, $i$ and $j$ are the element coordinates, $k_{p,q}$ is the convolution kernel, $x_{i+p-1, j+q-1}$ is the input data elements on the row $i+p-1$ and column $j+p-1$.

In terms of the selection of activation function, the convolution layer selects the rectified linear unit (ReLU) as an activation function, which is characterized by fast convergence and simple gradient calculation. Its expression shows that:

$$h(y_t) = \begin{cases} y_t & (y_t > 0) \\ 0 & (y_t \leq 0) \end{cases} \quad (13)$$

Also, the number of network-free parameters is reduced because the neurons on a mapping surface share the weights. Each convolution layer in the CNN is followed by a computational layer for local average and secondary extraction. As a result, this unique two-feature extraction structure reduces the feature resolution.

3.1.2. Pooling layer

The main function of the pooling layer is to divide the input data into multiple non-overlapping areas and to take the maximum value (maximum pooling) or average value (average pooling) from the values of each area. The expressions show that:

$$y' = \max_{i,j \in R}(y_{i,j}) \tag{14}$$

$$y' = \frac{1}{S_R} \sum_{i,j \in R} (y_{i,j}) \tag{15}$$

where $R$ is the pool region, $S_R$ is the pool area. The non-critical feature samples are eliminated to reduce the number of parameters by the pooling process. Thus, the training efficiency and estimation accuracy are improved, which is a critical part of data features extraction.

3.1.3. Full connection layer

The main function of the full connection layer is to expand the two-dimensional output data through the convolution layer and the pooling layer into one-dimensional data. Thereafter, the mapping relationship is obtained between the key characteristics of each factor and the industrial load value.

$$z = f(\sum_{k}^{K} w_k d_k + b) \tag{16}$$

where $d_k$ is the k-dimension input variable, $w$ is the corresponding weight, $b$ is the offset, and $z$ is the output.

*3.2. SRU data training model*

At present, the non-invasive decomposition of high energy-consuming industrial load faces some challenges. One of them is the difficulty in data collection, the other is the high dimension of dataset which requires a long calculation time and may affect the calculation accuracy. The traditional deep learning methods depend on the previous state of the input data when processing time series heavily, which increases the calculation difficulty and extends the calculation time. It is shown that SRU can fully retain data characteristics when processing medium and long-term data among the existing research results. Therefore, the SRU deep learning network is selected in this paper to train the high energy-consuming industrial load data. Also, a double-layer structure combined with CNN is established to decompose the non-invasive IPL data. The SRU trains input data through a chain structure of repetitive modules which can avoid the problem of long-term dependence relying on the internal unique design. The structure of the SRU unit is shown in Fig. 6, where each module contains an input gate, a forgetting gate, and an output gate (Hossain et al.,2021).

The information discarded is determined through the forgetting gate, that is:

$$S_t = \begin{bmatrix} x_{IPL} \\ z \end{bmatrix} \tag{17}$$

$$\tilde{s}_t = W s_t \tag{18}$$

$$f_t = \sigma(W_f x_t + b_f) \tag{19}$$

where $S_t$ is the high energy-consuming industrial load data and the eigenvalue matrix output by the CNN. $f_t$ is the state of forgetting gate, $W_f$ and $b_f$ are the weight and bias of forgetting gate, respectively, $\sigma$ is the activation function of the sigmoid layer. It can be seen from formula (19) that the SRU discards the intermediate output state, $h_{t-1}$, compared with the traditional LSTM. Hence, a parallel calculation is carried out to reduce calculation time.

The information that needs to be updated is determined through the input gate. Then, a new candidate vector is created by a Tanh module. However, in the process of load data decomposition, the accuracy of decomposition may not be that high because of the large amount of calculation, the disappearance of the gradient, slow convergence speed, and other problems. Therefore, the ReLU function is used as an activation function.

$$r_t = \sigma(W_r s_t + b_r) \tag{20}$$

$$c_t = f_t * c_{t-1} + (1 - f_t) * \tilde{s}_t \tag{21}$$

where $r_t$ is the state of the input gate, $W_r$ and $b_r$ are the weight and offset of the input gate, respectively, in equation (20), and $c_t$ is the state of the memory unit in equation (21).

The information to be output for updating the cell status is determined. Through the cooperation between the sigmoid layer and ReLU layer, the output information is determined as:

$$h_{il\ t} = r_t * Relu(c_t) + (1 - r_t) * s_t \tag{22}$$

where $h_{il\ t}$ is the final output which is the decomposition result of high energy-consuming industrial load data. ReLU function makes the network sparse, alleviates the occurrence of overfitting problems, and excavates the sample characteristics in-depth. The gradient of the ReLU function is constant in a non-negative interval, so the gradient does not disappear. This makes the model converge faster.

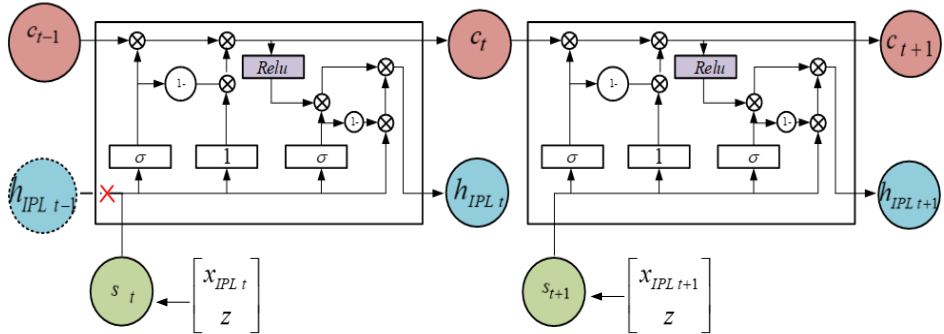

**Fig. 6.** SRU structure diagram

## 4. Design Framework and the Algorithm Flow

The decomposition system for the industrial park aggregation model including data acquisition and data preprocessing is proposed in this paper based on the NILM algorithm. The basic design framework is shown in Fig. 7:

(1) Data acquisition module: The daily load data of each factory in the industrial park is recorded by the smart meters and transmitted to the data center of the load aggregator through the network which can provide data support for sub-application modules.

(2) Data noise reduction module: Data processing is required as the data collected by smart meters usually contains a large amount of noise, especially for the distribution network which contains flexible load and dispatchable loads, such as, electrical appliances with inverter and distributed generation of renewable energy. The improved VMD algorithm is proposed to decompose the industrial load data into several modes with different frequencies. Then, the modes are reconstructed to reduce the raw IPL data noise. The advantage of this algorithm is that the optimal K value can be determined automatically based on the mean curvature of the modal instantaneous frequency to weaken the uncertainty influence of the empiric parameter set.

(3) Characteristic extraction module: Unlike the traditional NILD algorithm which depends on the relationship between the load variation tendency and time to decompose the data, a data feature extraction module is added innovatively. The extracted data characteristics and industrial load data are input into the training module together to improve the decomposition accuracy.

(4) Decomposition training module: Training decomposition module is constructed by SRU Compared with LSTM and GRU modules, SRU can achieve parallel training without dependence on the state of the previous moment. Therefore, it can greatly improve the training accuracy and achieve accurate decomposition.

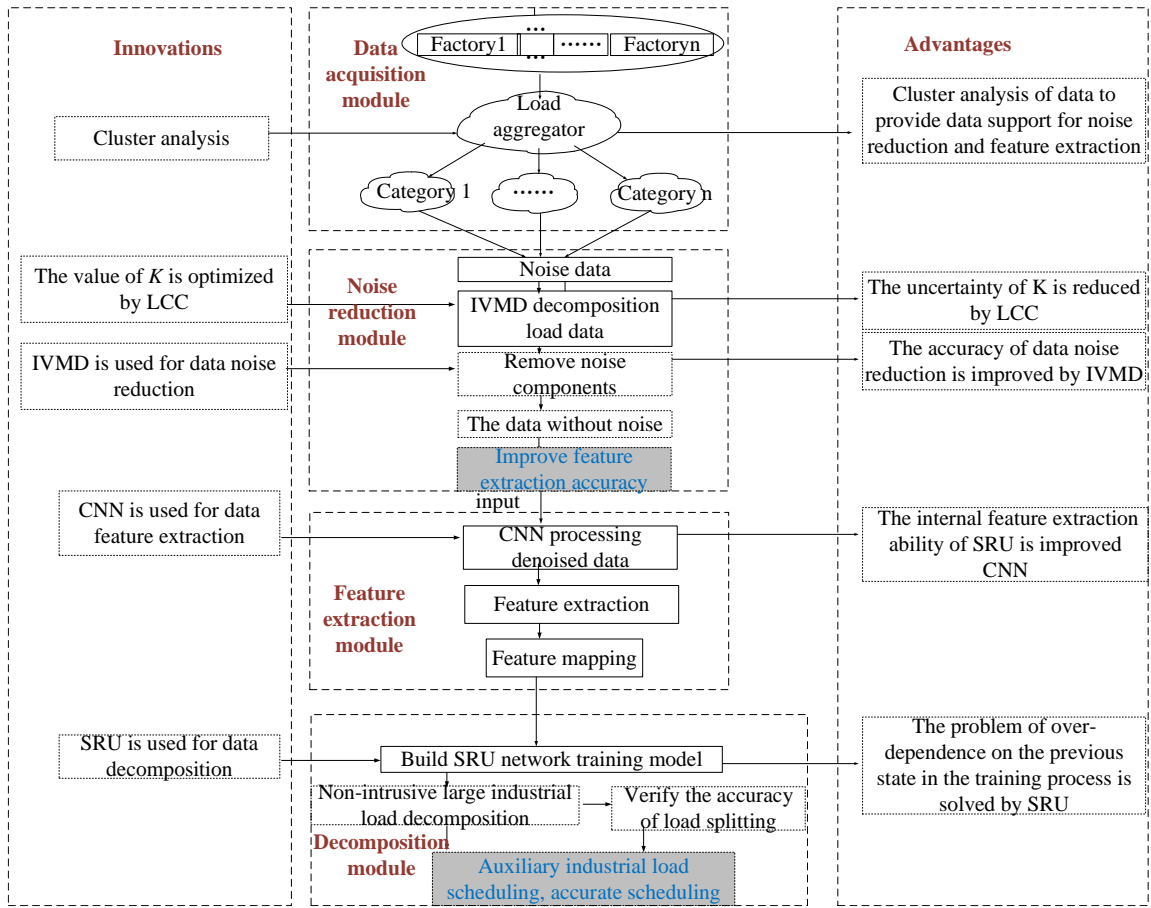

**Fig. 7.** Basic design framework and algorithm flow

## 5. Numerical Results

In this section, based on the load data of an industrial park in Liaoning Province in 2019, IVMD-CNN-SRU is used for case study to determine its effectiveness. The data interval is 15 minutes, and 96 points per day. The industrial park contains 41 enterprises. To prove the effectiveness of the proposed method, the relevant comparisons are made. Note that, this method can be widely applied to any region of the world and we simply use this case in China to demonstrate its implementation.

*5.1. The algorithm parameter selection and simulation settings*

For its real applications, the IVMD algorithm is used to eliminate the noise and discrete points from the sampled data to improve the training accuracy. At the same time, a part of the sampled REDD dataset which has the same sampling period as auxiliary training is used to improve the accuracy of the decomposition results. The clustering algorithm is used to process the sample data, and the clustering centers of four main loads are obtained. The detailed data profiles are shown in the following table:

**Table 2.** Clustering result statistics

| Category | Category capacity (kW) | Capacity ratio | Number of categories | Proportion of quantity |
|---|---|---|---|---|
| 1 | 1534.2513 | 0.2070 | 13 | 0.3173 |
| 2 | 4328.5157 | 0.5840 | 15 | 0.3658 |
| 3 | 758.9726 | 0.1024 | 7 | 0.1707 |
| 4 | 790.1025 | 0.1066 | 6 | 0.1462 |
| Total | 7411.8421 | 1 | 41 | 1 |

To eliminate the influence of the large difference in the load data amplitude, the data is normalized during clustering. It can be seen from Table 2 that the first type and the second type account for the largest proportion in the clustering result, and they are representatives of the IPL. Hence, the load data of the first-type textile and second-type building material enterprise are selected for simulation.

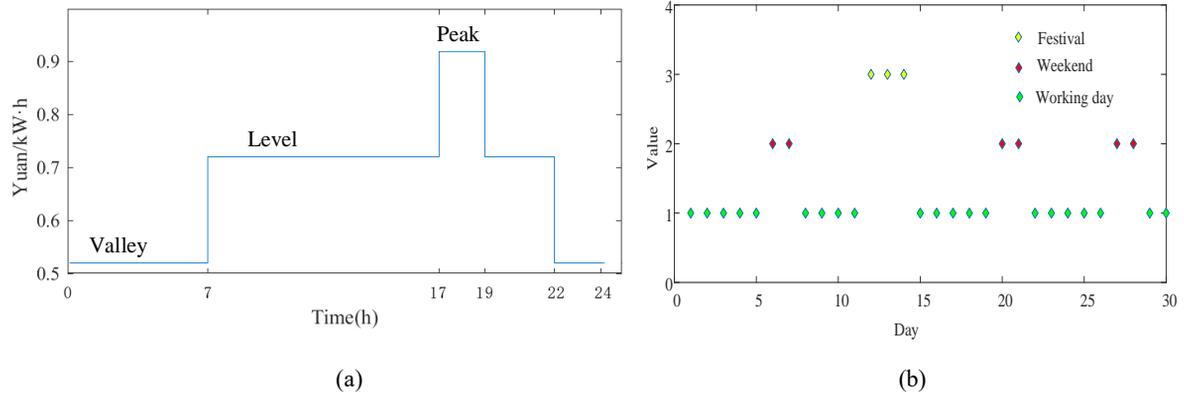

**Fig. 8.** Electricity price and Calendar effect (a) The time-sharing industrial electricity price in Liaoning (b) Calendar effect

The time-dependent electricity price of the industrial park in Liaoning Province is shown in Fig. 8 (a). It can be seen that industrial electricity price is the cheapest in the valley period and higher in the flat period. The highest electricity price in the peak period of electricity consumption is 0.9182. Fig.8 (b) shows the calendar effect in which the value of the festival is set 3, the value of the weekend is set 2, and the value of the working day is set 1.

The hardware environment of this paper is Intel (R) CoreTM i7-4710MQ CPU @ 2.5GHz, 8G DDR4 memory, and the software platform is WINDOWS-7 (64-bit) flagship operating system. CNN-SRU double-layer training network is built by Pycharm 3.2.0 and TensorFlow 2.0 deep learning framework, the detail parameters are shown in Table 3 and Table 4.

**Table 3.** Structure of convolution neural network

| Type | Channel number | Convolution kernel / moving step | Activation function | Pool type |
|---|---|---|---|---|
| Input | 1 | -- | -- | -- |
| Convolution layer | 16 | 5*5/1 | ReLU | -- |
| Pooling layer | 16 | 2*2/2 | -- | Max pooling |
| Fully connected layer | 128 | -- | -- | -- |
| Discard | -- | -- | -- | -- |
| Classifier | 4 | -- | -- | -- |

**Table 4.** Structure of simple recurrent unit

| Type | Layer number | Cell number |
|---|---|---|
| Convolution layer | 1 | 16 |
| Forward SRU | 1 | 64 |
| Reverse SRU | 1 | 64 |
| Fully connected layer | 2 | 128 |

*5.2. Simulation decomposition results with different algorithms*

The measured data above is set as input into the CNN-SRU network to be trained. Then, the effectiveness of the proposed algorithm is verified by testing the dataset. To reflect the advantages of the algorithm proposed in this paper in the non-invasive load decomposition of the high energy-consuming industrial loads, other existing methods are used to train the same dataset and the training results are tested. The comparison methods are ALIP (Bhotto et al., 2017) and attention-seq2seq (Wang et al., 2019). The load data of typical enterprises within one month is selected as the training datasets for training respectively. The training results are shown in Fig. 9 and Fig. 10. To make the decomposition results clearer, only a typical part operation results are shown in the figure.

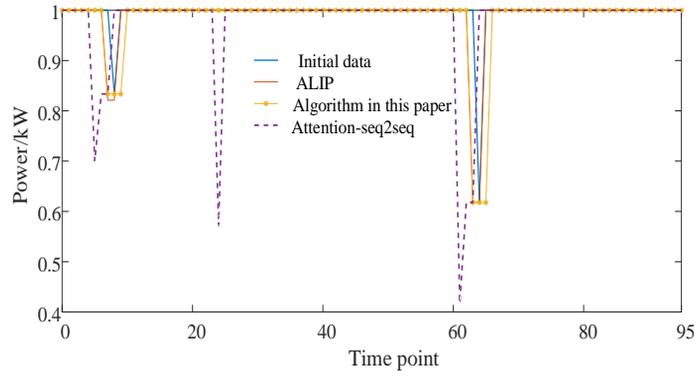

(a)

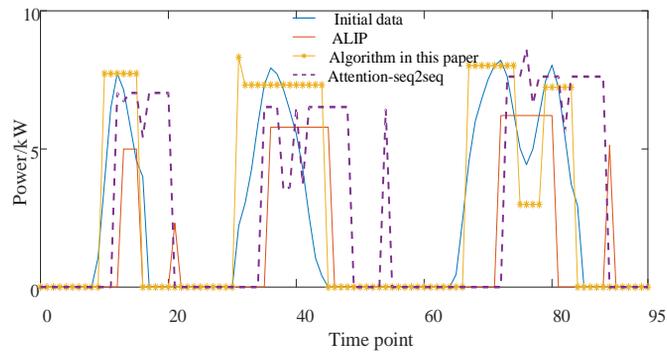

(b)

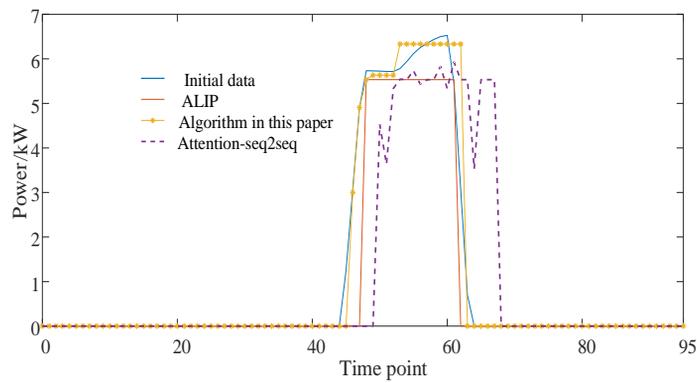

(c)

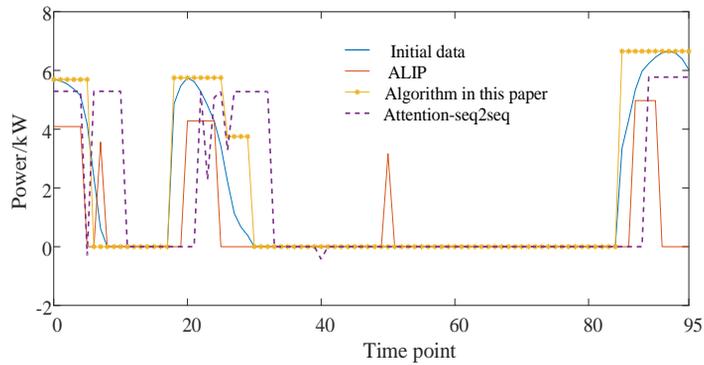

(d)

**Fig. 9.** Decomposition results of textile enterprise (a) Auxiliary power (b) Equipment 1 (c) Equipment 2 (d) Equipment 3

It can be seen from Fig. 9 that the proposed method is superior to the traditional algorithm in terms of the load decomposition accuracy and the switching state of industrial equipment. Hence, it has a great advantage in the non-invasive decomposition of high energy-consuming industrial loads. Taking the decomposition result of equipment 1 as an example, the equipment is disconnected during the time point 60-95. The disconnection time displayed by the decomposition results of the ALIP and attention-seq2seq algorithm is quite different from the actual situation. The decomposition result of the proposed algorithm in this paper not only predicts the disconnection time of the equipment but also retains the operating characteristics of the equipment to the maximum extent. The main reason that the algorithm is based on the data characteristics. It is different from the traditional algorithm which relies too much on the time stamp, which results in the decomposition result only approaching the real data in amplitude. But the decomposition results show an overall drift due to the influence of the previous IPL data state. At the same time, the SRU network unit does not need the previous state as input data during the training, which reduces the possibility of overall drift in the decomposition results.

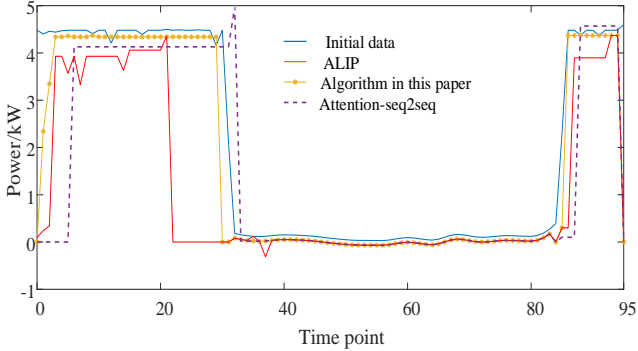

(a)

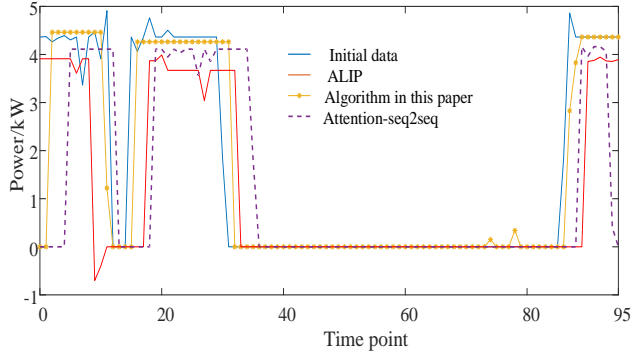

(b)

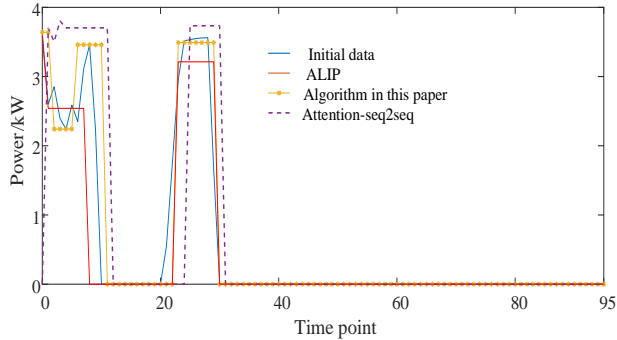

(c)

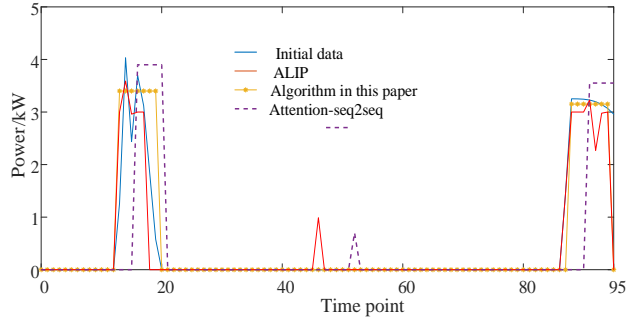

(d)

**Fig. 10.** Decomposition results of building materials enterprise (a) Equipment 1 (b) Equipment 2 (c) Equipment 3 (d) Equipment 4

It can be seen from Fig. 10 that the load data of building materials enterprise are different from that of textile enterprise. Their operation time is different and the equipment power of building materials enterprise is significantly higher than that of textile enterprise. This is due to its less equipment disconnection caused by its characteristics in which only one or two disconnection occurs in one operation cycle. The factors of different equipment types and operation modes have a great impact on the non-invasive IPL decomposition. Conventional decomposition methods decompose the fixed equipment in the case of the family and commercial. However, they have low accuracy when dealing with the problem of high energy consumption of industrial load decomposition. The algorithm proposed in this paper introduces data characteristics innovatively as the input of the training network which results in a simplified training process and improved decomposition accuracy.

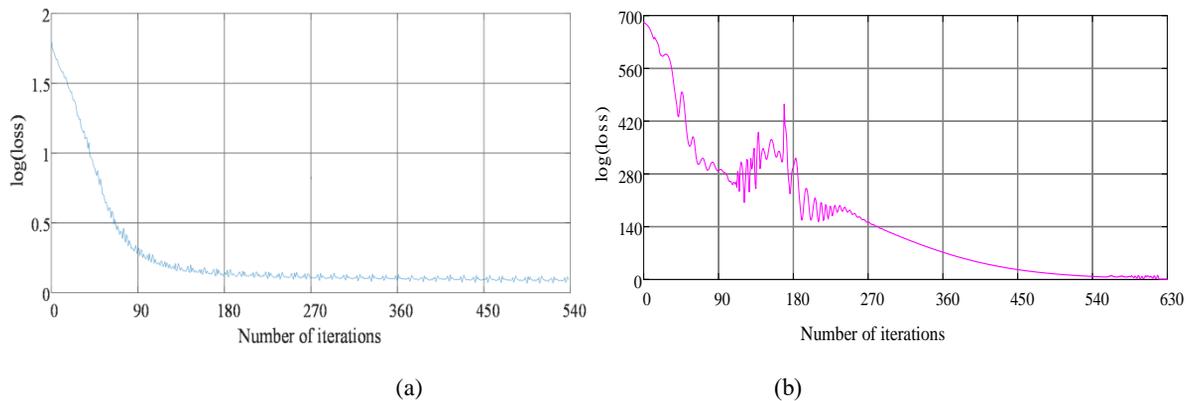

(a)                      (b)

**Fig. 11.** CNN and SRU partial operation results (a) CNN function loss (b) SRU function loss

From the loss function diagram of the CNN and SRU training networks, it can be seen that both of them have achieved the expected effect after 540 and 630 iterations, respectively. Also, the loss function gradually stabilizes without a change, which shows that the established CNN and SRU networks can converge quickly and are effective in dealing with the IPL data.

*5.3. Evaluation of simulation results with different algorithms*

To avoid the influence of subjective factors, four important judgment indices of NILM are introduced. These include accuracy, precision, recall, and F1 score (Wang et al.,2020).

$$A_{cc} = \frac{TP + TN}{P + N} \tag{23}$$

$$Precision = \frac{TP}{TP + FP} \tag{24}$$

$$Recall = \frac{TP}{TP + FN} \tag{25}$$

$$F1 = \frac{Precision * Recall}{Precision + Recall} \tag{26}$$

In equations (23)-(26), true positive (TP) is determined to be a positive class if the true value is a positive class; false positive (FP) is determined to be positive if the real value is negative; true negative (TN) is determined to be negative if the real value is negative; false negative (FN) is determined to be negative if the real value is positive.

Four evaluation indices are used to analyze the accuracy of the simulation results in section 3. Then, the decomposition accuracy of the algorithm proposed in this paper is compared with that of the other four existing non-invasive load decomposition algorithms. The selected index is the mean value of the evaluation indices of decomposition results under two cases. The results are shown in Table 5.

**Table 5.** Precision comparison of non-intrusive load decomposition algorithms

| Category | Index / Algorithm | Accuracy | Precision | Recall | F1 score |
|---|---|---|---|---|---|
| Textile enterprises | The proposed algorithm | 0.9896 | 0.9597 | 0.8947 | 0.4630 |
| | ALIP | 0.8532 | 0.8355 | 0.7409 | 0.3926 |
| | Attention-seq2seq | 0.9610 | 0.9483 | 0.8546 | 0.4495 |
| | IVMD-SRU | 0.8066 | 0.7409 | 0.7152 | 0.3673 |
| | CNN-SRU | 0.8647 | 0.8406 | 0.7923 | 0.4015 |
| Building materials enterprises | The proposed algorithm | 0.9586 | 0.9243 | 0.8495 | 0.4531 |
| | ALIP | 0.8371 | 0.8185 | 0.7191 | 0.3534 |
| | Attention-seq2seq | 0.9264 | 0.9130 | 0.8137 | 0.3843 |
| | IVMD-SRU | 0.8314 | 0.7948 | 0.7754 | 0.3675 |
| | CNN-SRU | 0.8759 | 0.8236 | 0.7876 | 0.4032 |

It can be seen from the comparison results of the algorithm decomposition accuracy in Table 5 that the proposed algorithm in this paper is superior to other algorithms in terms of decomposition accuracy. The GRU network model is used by attention-seq2seq to train the IPL data, which heavily depends on the state of the device in the previous period. However, the power consumption pattern of the equipment included in the high energy-consuming industrial load is not closely related to the time, which leads to the low accuracy of the final decomposition of the algorithm. The accuracy of IVMD-SRU and CNN-SRU are also low relatively.

Firstly, the accuracy of the algorithm proposed in this paper increased by 19.02% and 3.14% compared with ALIP and attention-seq2seq, which fully reflects the advantages of the deep learning algorithm in NILD. Therefore, the proposed algorithm in this paper has higher accuracy in high energy-consuming industrial load decomposition. Secondly, from the perspective of the precision of different algorithms, the precision of the algorithm proposed in this paper is 0.9597, which is 14.87%, 1.20%, 29.53%, and 14.17% higher than the accuracies of ALIP, attention-seq2seq, IVMD-SRU and CNN-SRU, respectively. The precision results indicate that the decomposition results of the proposed algorithm have higher prediction precision for positive values. Thirdly, regarding the recall rate of the simulation results, the recall rate of the proposed algorithm is 0.8947, which is 20.76%, 4.69%, 25.10%, and 12.92% higher than those of the other four algorithms, respectively. The simulation results show that the algorithm has the best decomposition recall rate, and it is more accurate to predict the positive value of the input data. Finally, the F1 score is a comprehensive evaluation index for the accuracy rate and recall rate. From the above analysis, it can be seen that the accuracy and recall rate of the proposed algorithm is better than the four comparison algorithms significantly, and the F1 score is also optimal. Compared with the other four algorithms, the F1 scores of the proposed algorithm are increased by 17.93%, 3.00%, 26.05%, and 15.32%, respectively. Therefore, the algorithm proposed in this paper fully extracts the characteristics of load curve decomposition process and improves the accuracy of positive result value prediction.

*5.4. Evaluation of simulation results with different industry categories*

In practice, another problem faced by industrial park aggregation model recognition is that the diversity of enterprise categories in the park leads to confusion in the pattern recognition. The training accuracy of different equipment in different enterprises is not enough, resulting in a decrease in the accuracy of decomposition. To tackle this problem, we verified the influence of the increase of enterprise categories in the industrial park on the accuracy of algorithm decomposition through simulation. Three cases are set for comparison. Case one, two, and three contain five, ten, and fifteen factories respectively. The results are shown in Fig. 12.

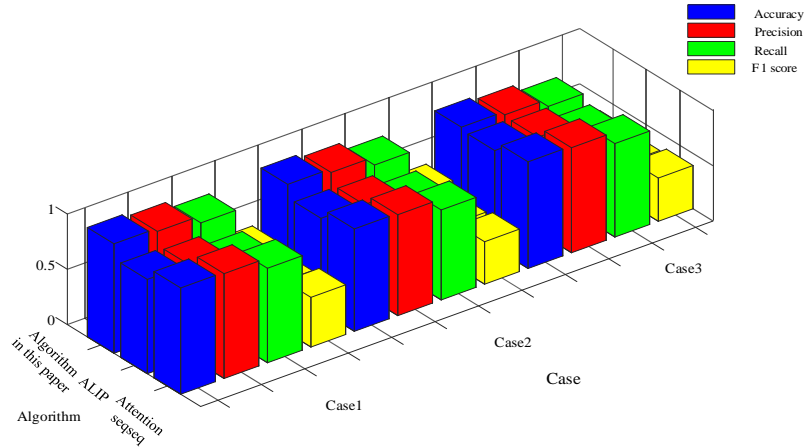

**Fig. 12.** Simulation error with different categories

It can be seen from Fig. 12, with the increase of plant type and internal equipment type, the decomposition accuracy of the algorithm is above 0.9. In addition, with the increase of factory categories, the recognition accuracy of the other two algorithms decreases significantly. This is mainly due to the innovative introduction of CNN network pooling data extraction characteristics in this paper, which better retains the characteristics of factory equipment load data rather than excessively relying on time.

The evaluation of big data processing methods is mainly reflected in two aspects: one is the accuracy of data processing, and the other is the speed of data processing. Only high-efficiency data computational ability could maximize the timeliness of the data. The running time of the methods is analyzed, and the results are shown in Table 6.

**Table 6.** Comparison of algorithm running time

| Algorithm | Running time |
| --- | --- |
| The proposed algorithm | 90.70s |
| ALIP | 159.21s |
| Attention-seq2seq | 206.75s |

From the data in the table, it can be seen that under the same hardware conditions, the algorithm only takes 90.70 seconds. Compared with ALIP and attention-seq2seq, the operation time is reduced by 43.03% and 56.13% respectively. Because the SRU training module can perform parallel operation on the data and does not rely on the previous state data, it can ensure the data training accuracy and improve the training speed. In a word, the proposed algorithm has obvious advantages over the existing two algorithms.

## 6. Conclusions

An improved VMD algorithm combined with the double-layer deep learning model is proposed to build a multi-functional non-invasive decomposition framework for high energy-consuming industrial park loads which contains IPL data denoising and decomposition. The results show that:

(1) The K value setting method, based on the instantaneous frequency mean quantization analysis, is able to determine the decomposition mode number K accurately. This method improves the instability of the traditional parameter optimization algorithm and the subjectivity of human empirical judgment. The bad points and noises in the raw data are removed effectively by IVMD processing, which improves the data utilization rate by 19.49%.

(2) Through the CNN feature extraction, the internal feature extraction ability of SRU is improved. Moreover, the problem of over-dependence on the previous state in the training process of the conventional decomposition algorithm is solved resulting in a higher decomposition accuracy of IPL data. The accuracy of the algorithm proposed in this paper increases by 19.02% and 3.14% compared with ALIP and Attention seq2seq, respectively. The F1 score of the algorithm increases by 17.93% and 3.00% compared with ALIP and Attention seq2seq, respectively.

(3) The double-layer deep learning model is proposed in this paper. Because of the parallelized calculation of the SRU algorithm, the operation time of the proposed algorithm is reduced by 43.03% and 56.13% compared with ALIP and attention-seq2seq, respectively. The results show that proposed algorithm is more accurate and effective in dealing with a non-invasive decomposition of high energy-consuming industrial loads.

## Acknowledgment

This work is supported in part by the scientific research project of education department of Liaoning province (LJKZ0129) and science and technology project of state grid corporation of China (SGTYHT/21-JS-221)

## References

Arnold, B., Lutz, T., Krämer, E., et al, 2018. Wind-turbine trailing-edge noise reduction by means of boundary-layer suction. Aiaa Journal. 56, 1-12. https://doi.org/ 10.2514/1.J056633.

Azizi, E., Ahmadiahangar, R., Rosina, A., et al, 2021. Residential energy flexibility characterization using non-intrusive load monitoring. Sustainable Cities and Society. 103321, 2210-6707. https://doi.org/10.1016/j.scs.2021.103321.

Bhotto, M. Z. A., Makonin, S. and Bajić, I. V., 2016. Load Disaggregation Based on Aided Linear Integer Programming. IEEE Transactions on Circuits and Systems II: Express Briefs. 64 (7), 792-796, https://doi.org/10.1109/TCSII.2016.2603479.

Brucke, K., Arens, S., Telle, J.-S., et al, 2021. A non-intrusive load monitoring approach for very short-term power predictions in commercial buildings. Applied Energy. 292(116860), 0306-2619. https://doi.org/10.1016/j.apenergy.2021.116860.

Ch'ien, L.-B., Wang, Y.-J., Shi, A.-C., et al, 2020. Noise suppression: Empirical modal decomposition in non-dispersive infrared gas detection systems. Infrared Physics & Technology. 108 (103335), 1350-4495. https://doi.org/10.1016/j.infrared.2020.103335.

Cui, H., Zhou, K., 2018. Industrial power load scheduling considering demand response. Journal of Cleaner Production. 204 (447-460), 0959-6526. https://doi.org/10.1016/j.jclepro.2018.08.270.

Cui, J., Pan, J., et al, 2022. Improved normal-boundary intersection algorithm: A method for energy optimization strategy in smart buildings. Building and Environment. 212 (108846), 0360-1323. https://doi.org/10.1016/j.buildenv.2022.108846.

Cui, J., Yu, R., et al, 2019. Intelligent load pattern modeling and denoising using improved variational mode decomposition for various calendar periods. Applied Energy. 247, 480-491. https://doi.org/10.1016/j.apenergy.2019.03.163.

Dragomiretskiy, K., Zosso, D., 2014. Variational mode decomposition. IEEE Transactions on Signal Processing, 62, 531-544. https://doi.org/10.1109/TSP.2013.2288675.

Golpîra, H., Bahramara, S., 2020. Internet-of-things-based optimal smart city energy management considering shiftable loads and energy storage. Journal of Cleaner Production. 264(121620), 0959-6526. https://doi.org/10.1016/j.jclepro.2021.126564.

Hossain, M. A., Chakrabortty, R. K., Elsawah, S., et al, 2021. Very short-term forecasting of wind power generation using hybrid deep learning model. Journal of Cleaner Production. 296 (126564), 0959-6526. https://doi.org/10.1016/j.jclepro.2021.126564.

Meier, A., Cautley, D., 2021. Practical limits to the use of non-intrusive load monitoring in commercial buildings. Energy and Buildings. 251(111308), 0378-7788. https://doi.org/10.1016/j.enbuild.2021.111308.

Qureshi, M., Ghiaus, C., Ahmad, N., 2021. A blind event-based learning algorithm for non-intrusive load disaggregation. International Journal of Electrical Power & Energy Systems. 129 (106834), 0142-0615. https://doi.org/10.1016/j.ijepes.2021.106834.

Rana, A., Perera, P., Ruparathna, R., Karunathilake, H., et al, 2020. Occupant-based energy upgrades selection for Canadian residential buildings based on field energy data and calibrated simulations. Journal of Cleaner Production. 271 (122430), 0959-6526. https://doi.org/10.1016/j.jclepro.2020.122430.

Sharma, M., Sharma, P., Pachori, R.B., et al, 2018. Dual-tree complex wavelet transform-based features for automated alcoholism identification. International Journal of Fuzzy Systems. 20, 1297-1308, https://doi.org/10.1007/s40815-018-0455-x.

Subash, R. K., Jayashree, E., et al, 2021. Non -intrusive load monitoring technique using deep neural networks for energy disaggregation. Materials Today: Proceedings. 2214-7853. https://doi.org/10.1016/j.matpr.2021.01.192.

Takada, K. and Satoh, S. I., 2018. Beat noise reduction utilizing the transient acoustic-wave response of an optical fiber in Brillouin grating-based optical low coherence reflectometry. Applied Optics, 57, 5235. https://doi.org/10.1364/AO.57.005235.

Verma, S., Singh, S., Majumdar, A., 2021. Multi-label LSTM autoencoder for non-intrusive appliance load monitoring. Electric Power Systems Research. 199(107414), 0378-7796. https://doi.org/10.1016/j.epsr.2021.107414.

Wang, A. L.; Chen, B. X.; Wang, C. G. et al, 2018. Non-intrusive load monitoring algorithm based on features of V–I trajectory. Electric Power Systems Research. 157, 134-144. https://doi.org/10.1016/j.epsr.2017.12.012.

Wang, K., Zhong, H., Yu, N., Xia, Q., 2019. Nonintrusive load monitoring based on sequence-to-sequence model with attention mechanism. Proceedings of The Chinese Society for Electrical Engineering. 39 (01), 75-83. https://doi.org/10.13334/j.0258-8013.pcsee.181123.

Wang, R., Asghari, V., Hsu, S. C., Lee, C. J., & Chen, J. H., 2020. Detecting corporate misconduct through random forest in china's construction industry. Journal of Cleaner Production. 268, 122266. https://doi.org/10.1016/j.jclepro.2020.122266.

Wu, Z., Chao, W., Peng, W. X., Liu, W. H., Zhang, H. Q., 2021. Non-intrusive load monitoring using factorial hidden markov model based on adaptive density peak clustering. Energy and Buildings. 244(111025), 0378-7788. https://doi.org/10.101.

Xie, M., Zhou, K., Huang, Y. L., et al, 2019．A white noise suppression method for partial discharge based on short time singular value decomposition. Proceedings of the CSEE. 39(3), 915-922. https://doi.org/10.13334/j.0258-8013.pcsee.172747.

Zhou, X., Feng, J., Li, Y., 2021, Non-intrusive load decomposition based on CNN–LSTM hybrid deep learning model. Energy Reports. 7, 5762-5771, 2352-4847. https://doi.org/10.1016/j.egyr.2021.09.001.